\def\year{2018}\relax
\documentclass[letterpaper]{article} %DO NOT CHANGE THIS
\usepackage{aaai18}  %Required
\usepackage{times}  %Required
\usepackage{helvet}  %Required
\usepackage{courier}  %Required
\usepackage{url}  %Required
\usepackage{graphicx}  %Required
\usepackage{subcaption}
\usepackage{tikz}

\usepackage{pgfplots}
\pgfplotsset{compat=1.14}
\usepackage{amsmath, amssymb}
\usepackage{booktabs}

\newcommand{\citet}[1]{\citeauthor{#1}~\shortcite{#1}}
\newcommand{\citep}{\cite}

\nocopyright

\begin{document}

\title{Modeling Relational Data with Graph Convolutional Networks}

\author{
  Michael Schlichtkrull$^*$\\
  University of Amsterdam\\
  \texttt{m.s.schlichtkrull@uva.nl}\\\And
  Thomas N.~Kipf\thanks{Equal contribution.}\\
  University of Amsterdam\\
  \texttt{t.n.kipf@uva.nl}\\ \And
   Peter Bloem\\
  VU Amsterdam\\
  \texttt{p.bloem@vu.nl}\\ \AND
  Rianne van den Berg\\
  University of Amsterdam\\
  \texttt{r.vandenberg@uva.nl}\\\And
  Ivan Titov\\
  University of Amsterdam\\
  \texttt{titov@uva.nl}\\ \And
  Max Welling\\
  University of Amsterdam, CIFAR\thanks{Canadian Institute for Advanced Research}\\
  \texttt{m.welling@uva.nl}\\
}

\maketitle

\begin{abstract}

Knowledge graphs enable a wide variety of applications, including question answering and information retrieval. Despite the great effort invested in their creation and maintenance, even the largest (e.g., Yago, DBPedia or Wikidata) remain incomplete. We introduce Relational Graph Convolutional Networks (R-GCNs) and apply them to two standard knowledge base completion tasks: Link prediction (recovery of missing facts, i.e.~subject-predicate-object triples) and entity classification (recovery of missing entity attributes). R-GCNs are related to a recent class of neural networks operating on graphs, and are developed specifically to deal with the highly multi-relational data characteristic of realistic knowledge bases. We demonstrate the effectiveness of R-GCNs as a stand-alone model for entity classification. We further show that factorization models for link prediction such as DistMult can be significantly improved by enriching them with an encoder model to accumulate evidence over multiple inference steps in the relational graph, demonstrating a large improvement of 29.8\% on FB15k-237 over a decoder-only baseline.

\end{abstract}

\section{Introduction}
Knowledge bases organize and store factual knowledge, enabling
a multitude of applications including question answering~\cite{yao2014information,bao2014knowledge,seyler2015generating,hixon2015learning,bordes2015large,dong2015question} and information retrieval~\cite{kotov2012tapping,dalton2014entity,xiong2015query,xiong2015esdrank}.
Even the largest knowledge
bases (e.g. DBPedia, Wikidata or Yago), despite enormous effort invested in their maintenance, are incomplete, and the lack of coverage
harms downstream applications. Predicting missing
information in knowledge bases is the main focus of statistical relational learning (SRL).

\begin{figure}[t!]
    \centering
    \includegraphics[width=0.95\linewidth]{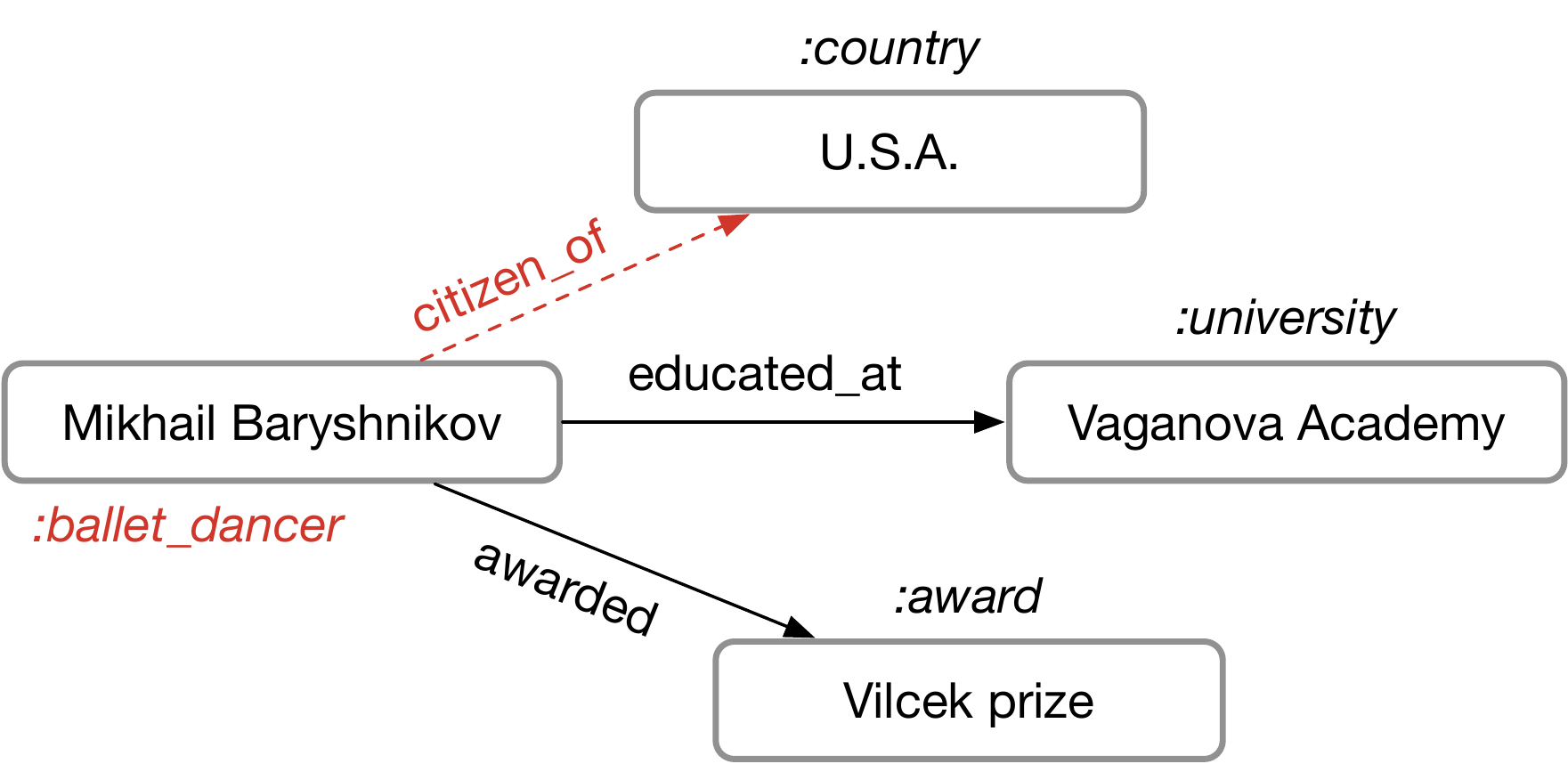}
    \caption{A knowledge base fragment: The nodes are entities, the edges are relations labeled with their types, the nodes are labeled with entity types (e.g., {\it university}). The edge and the node label shown in red are the missing information to be inferred.}
    \label{fig:kb}
\end{figure}

Following previous work on SRL, we assume that knowledge bases store collections of triples of the form (subject, predicate, object). Consider, for example, the triple (\textit{Mikhail Baryshnikov}, \textit{educated\_at}, \textit{Vaganova Academy}), where we will refer to {\it Baryshnikov} and {\it Vaganova Academy} as entities and to {\it educated\_at} as a relation. Additionally, we assume that entities are labeled with types (e.g.,  {\it Vaganova Academy} is marked as a {\it university}).
It is convenient to represent knowledge bases as directed labeled multigraphs with entities corresponding to nodes and triples encoded by labeled edges (see Figure~\ref{fig:kb}).

We consider two fundamental SRL tasks: link prediction (recovery of missing triples) and entity classification (assigning types or categorical properties to entities). In both cases, many missing pieces of information can be expected to reside within the graph encoded through the neighborhood structure -- i.e.  knowing that \textit{Mikhail Baryshnikov} was educated at the \textit{Vaganova Academy} implies both that \textit{Mikhail Baryshnikov} should have the label person, and that the triple (\textit{Mikhail Baryshnikov}, \textit{lived\_in}, \textit{Russia}) must belong to the knowledge graph. Following this intuition, we develop an encoder model for entities in the relational graph and apply it to both tasks.

Our entity classification model, similarly to \citet{kipf2016semi}, uses softmax classifiers at each node in the graph. The classifiers take node representations supplied by a relational graph convolutional network (R-GCN) and predict the labels. The model, including R-GCN parameters, is learned by optimizing the cross-entropy loss.

Our link prediction model can be regarded as an autoencoder consisting of (1) an encoder: an R-GCN producing latent feature representations of entities, and (2) a decoder: a tensor factorization model exploiting these representations to predict labeled edges. Though in principle the decoder can rely on any type of factorization (or generally any scoring function), we use one of the simplest and most effective factorization methods: DistMult~\cite{distmult-embedding_entities_and_relations}. We observe that our method achieves competitive results on standard benchmarks, outperforming, among other baselines, direct optimization of the factorization (i.e.~vanilla DistMult).  This improvement is especially large when we consider the more challenging FB15k-237 dataset~\cite{toutanova2015observed}. This result demonstrates that explicit modeling of neighborhoods in R-GCNs is beneficial for recovering missing facts in knowledge bases.

Our main contributions are as follows. To the best of our knowledge, we are the first to show that the GCN framework can be applied to modeling relational data, specifically to link prediction and entity classification tasks. Secondly, we introduce techniques for parameter sharing and to enforce sparsity constraints, and use them to apply R-GCNs to multigraphs with large numbers of relations. Lastly, we show that the performance of factorization models, at the example of DistMult, can be significantly improved by enriching them with an encoder model that performs multiple steps of information propagation in the relational graph.

\section{Neural relational modeling}
We introduce the following notation: we denote directed and labeled multi-graphs as $G = (\mathcal{V}, \mathcal{E}, \mathcal{R})$ with nodes (entities) $v_i \in \mathcal{V}$ and labeled edges (relations) $(v_i, r, v_j) \in \mathcal{E}$, where $r\in\mathcal{R}$ is a relation type.\footnote{$\mathcal{R}$ contains relations both in canonical direction (e.g. \textit{born\_in}) and in inverse direction (e.g. \textit{born\_in\_inv}).}

\subsection{Relational graph convolutional networks}
Our model is primarily motivated as an extension of GCNs that operate on local graph neighborhoods \cite{duvenaud2015convolutional,kipf2016semi} to large-scale relational data. These and related methods such as graph neural networks \cite{scarselli2009graph} can be understood as special cases of a simple differentiable message-passing framework \cite{gilmer2017neural}:
\begin{equation}
\label{eq:message-passing}
h_i^{(l+1)}= \sigma \left( \sum_{m \in \mathcal{M}_i} g_m(h_i^{(l)}, h_j^{(l)}) \right),
\end{equation}
where $h_i^{(l)}\in\mathbb{R}^{d^{(l)}}$ is the hidden state of node $v_i$ in the $l$-th layer of the neural network, with $d^{(l)}$ being the dimensionality of this layer's representations. Incoming messages of the form $g_m(\cdot, \cdot)$ are accumulated and passed through an element-wise activation function $\sigma(\cdot)$, such as the $\mathrm{ReLU}(\cdot)=\max(0,\cdot)$.\footnote{Note that this represents a simplification of the message passing neural network proposed in \cite{gilmer2017neural} that suffices to include the aforementioned models as special cases.} $\mathcal{M}_i$ denotes the set of incoming messages for node $v_i$ and is often chosen to be identical to the set of incoming edges. $g_m(\cdot, \cdot)$ is typically chosen to be a (message-specific) neural network-like function or simply a linear transformation $g_m(h_i, h_j)=W h_j$ with a weight matrix $W$ such as in \citet{kipf2016semi}.

This type of transformation has been shown to be very effective at accumulating and encoding features from local, structured neighborhoods, and has led to significant improvements in areas such as graph classification \cite{duvenaud2015convolutional} and graph-based semi-supervised learning \cite{kipf2016semi}.

Motivated by these architectures, we define the following simple propagation model for calculating the forward-pass update of an entity or node denoted by $v_i$ in a relational (directed and labeled) multi-graph:
\begin{equation}
\label{eq:layer}
h_i^{(l+1)}= \sigma \left( \sum_{r \in \mathcal{R}}\sum_{j \in \mathcal{N}^r_i} \frac{1}{c_{i,r}}W_r^{(l)} h_j^{(l)} + W_0^{(l)}h_i^{(l)} \right),
\end{equation}
where $\mathcal{N}^r_i$ denotes the set of neighbor indices of node $i$ under relation $r\in\mathcal{R}$. $c_{i,r}$ is a problem-specific normalization constant that can either be learned or chosen in advance (such as $c_{i,r}=|\mathcal{N}^r_i|$).

Intuitively, (\ref{eq:layer}) accumulates transformed feature vectors of neighboring nodes through a normalized sum. Different from regular GCNs, we introduce relation-specific transformations, i.e. depending on the type and direction of an edge.
To ensure that the representation of a node at layer $l+1$ can also be informed by the corresponding representation at layer $l$, we add a single self-connection of a special relation type to each node in the data. Note that instead of simple linear message transformations, one could choose more flexible functions such as multi-layer neural networks (at the expense of computational efficiency). We leave this for future work.

\begin{figure}[t!]
        \centering
    \includegraphics[width=0.8\linewidth]{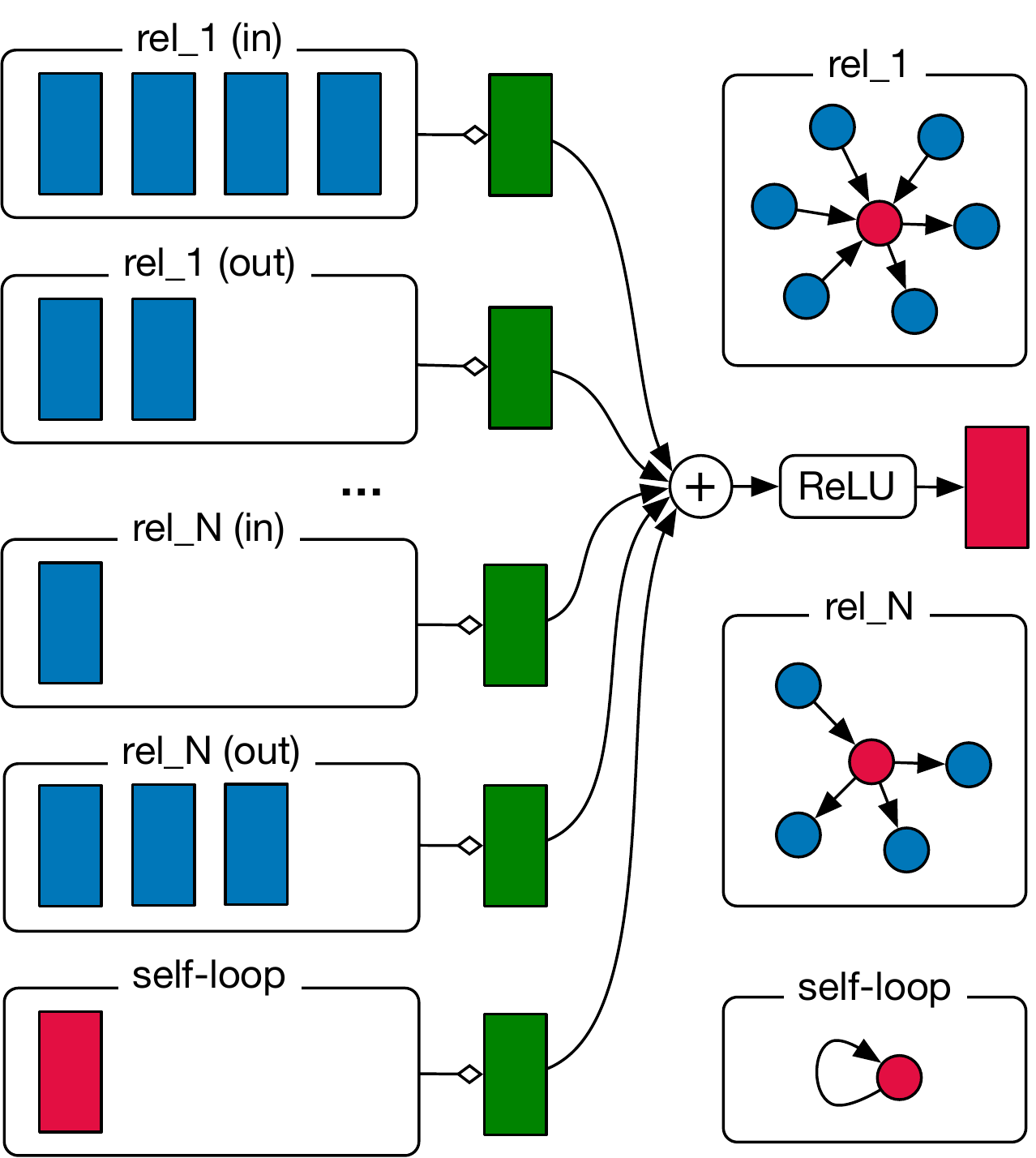}
    \label{fig:model-a}
    \caption{Diagram for computing the update of a single graph node/entity (red) in the R-GCN model. Activations ($d$-dimensional  vectors) from neighboring nodes (dark blue) are gathered and then transformed for each relation type individually (for both in- and outgoing edges). The resulting representation (green) is accumulated in a (normalized) sum and passed through an activation function (such as the ReLU). This per-node update can be computed in parallel with shared parameters across the whole graph.}
    \label{fig:model}
\end{figure}

A neural network layer update consists of evaluating (\ref{eq:layer}) in parallel for every node in the graph. In practice, (\ref{eq:layer}) can be implemented efficiently using sparse matrix multiplications to avoid explicit summation over neighborhoods. Multiple layers can be stacked to allow for dependencies across several relational steps. We refer to this graph encoder model as a relational graph convolutional network (R-GCN). The computation graph for a single node update in the R-GCN model is depicted in Figure \ref{fig:model}.

\subsection{Regularization}
A central issue with applying (\ref{eq:layer}) to highly multi-relational data is the rapid growth in number of parameters with the number of relations in the graph. In practice this can easily lead to overfitting on rare relations and to models of very large size.

To address this issue, we introduce two separate methods for regularizing the weights of R-GCN-layers: \textit{basis}- and \textit{block-diagonal}-decomposition. With the basis decomposition, each $W_r^{(l)}$ is defined as follows:
\begin{equation}
\label{eq:basis}
W_r^{(l)} = \sum_{b=1}^B a_{rb}^{(l)} V_b^{(l)},
\end{equation}
i.e.~as a linear combination of basis transformations $V_b^{(l)}\in\mathbb{R}^{d^{(l+1)}\times d^{(l)}}$ with coefficients $a_{rb}^{(l)}$ such that only the coefficients depend on \textit{r}. In the block-diagonal decomposition, we let each $W_r^{(l)}$ be defined through the direct sum over a set of low-dimensional matrices:
\begin{equation}
\label{eq:block}
W_r^{(l)} = \bigoplus_{b=1}^B Q^{(l)}_{br}.
\end{equation}
Thereby, $W_r^{(l)}$ are block-diagonal matrices:
$\mathrm{diag}(Q^{(l)}_{1r}, \ldots, Q^{(l)}_{Br})$ with $Q^{(l)}_{br} \in \mathbb{R}^{(d^{(l+1)}/B)\times( d^{(l)}/B)}$.

The basis function decomposition \eqref{eq:basis} can be seen as a form of effective weight sharing between different relation types, while the block decomposition \eqref{eq:block} can be seen as a sparsity constraint on the weight matrices for each relation type. The block decomposition structure encodes an intuition that latent features can be grouped into sets of variables which are more tightly coupled within groups than across groups.
Both decompositions reduce the number of parameters needed to learn for highly multi-relational data (such as realistic knowledge bases). At the same time, we expect that the basis parameterization can alleviate overfitting on rare relations, as parameter updates are shared between both rare and more frequent relations.

The overall R-GCN model then takes the following form: We stack $L$ layers as defined in \eqref{eq:layer} -- the output of the previous layer being the input to the next layer. The input to the first layer can be chosen as a unique one-hot vector for each node in the graph if no other features are present. For the block representation, we map this one-hot vector to a dense representation through a single linear transformation. While we only consider such a featureless approach in this work, we note that it was shown in \citet{kipf2016semi} that it is possible for this class of models to make use of pre-defined feature vectors (e.g.~a bag-of-words description of a document associated with a specific node).

\section{Entity classification}
\begin{figure}[t!]
    \centering
    \begin{subfigure}[b]{0.4\linewidth}
      \centering
      \includegraphics[width=0.90\linewidth, trim={0 0.2cm 0.6cm 0}, clip]{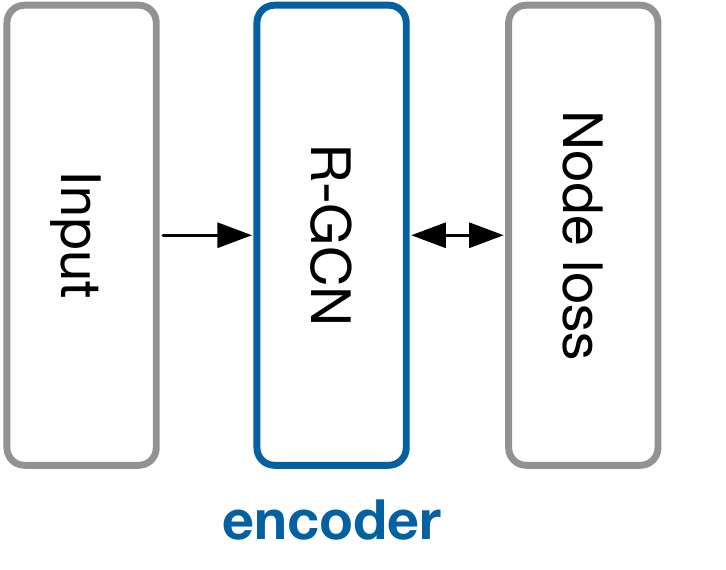}
      \caption{Entity classification}        
      \label{fig:model-b}
    \end{subfigure}%
    ~\quad
    \begin{subfigure}[b]{0.49\linewidth}
        \centering
        \includegraphics[width=\linewidth, trim={0 0.2cm 0.6cm 0}, clip]{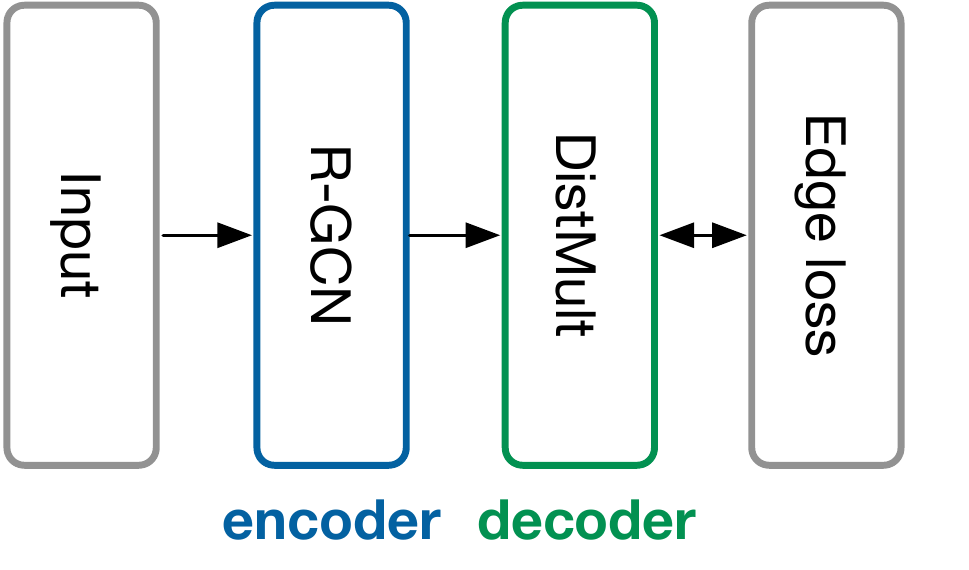}
        \caption{Link prediction}
        \label{fig:model-c}
    \end{subfigure}
    \caption{(a) Depiction of an R-GCN model for entity classification with a per-node loss function. (b) Link prediction model with an R-GCN encoder (interspersed with fully-connected/dense layers) and a DistMult decoder that takes pairs of hidden node representations and produces a score for every (potential) edge in the graph. The loss is evaluated per edge.}
\end{figure}

For (semi-)supervised classification of nodes (entities), we simply stack R-GCN layers of the form \eqref{eq:layer}, with a $\mathrm{softmax}(\cdot)$ activation (per node) on the output of the last layer. We minimize the following cross-entropy loss on all labeled nodes (while ignoring unlabeled nodes):
\begin{equation}
\mathcal{L}= -\sum_{i\in\mathcal{Y}}\sum_{k=1}^K t_{ik} \ln h_{ik}^{(L)}  \, ,
\label{eq:}
\end{equation} 
where $\mathcal{Y}$ is the set of node indices that have labels and $h_{ik}^{(L)}$ is the $k$-th entry of the network output for the $i$-th labeled node. $t_{ik}$ denotes its respective ground truth label. In practice, we train the model using (full-batch) gradient descent techniques. A schematic depiction of our entity classification model is given in Figure \ref{fig:model-b}.

\section{Link prediction}\label{section:link_prediction}
Link prediction deals with prediction of new facts (i.e. triples \textit{(subject, relation, object)}). Formally, the knowledge base is represented by a directed, labeled graph $G = (\mathcal{V},\mathcal{E},\mathcal{R})$. Rather than the full set of edges $\mathcal{E}$, we are given only an incomplete subset $\hat{\mathcal{E}}$. The task is to assign scores $f(s,r,o)$ to possible edges $(s,r,o)$ in order to determine how likely those edges are to belong to $\mathcal{E}$.

In order to tackle this problem, we introduce a graph auto-encoder model, comprised of an entity encoder and a scoring function (decoder). The encoder maps each entity $v_i \in \mathcal{V}$ to a real-valued vector $e_i \in \mathbb{R}^d$. The decoder reconstructs edges of the graph relying on the vertex representations; in other words, it scores \textit{(subject, relation, object)}-triples through a function $s: \mathbb{R}^d \times \mathcal{R} \times \mathbb{R}^d \to \mathbb{R}$. Most existing approaches to link prediction (for example, tensor and neural factorization methods~\cite{socher2013reasoning,lin2015modeling,toutanova2016compositional,distmult-embedding_entities_and_relations,complex-complex_embeddings_for_simple_link_prediction}) can be interpreted under this framework. The crucial distinguishing characteristic of our work is the reliance on an encoder. Whereas most previous approaches use a single, real-valued vector $e_i$ for every $v_i \in \mathcal{V}$ optimized directly in training, %(i.e.~using the identity function as an encoder model) -- IT I do not think it is clear at that stage
we compute representations through an R-GCN encoder with $e_i = h_i^{(L)}$, similar to the graph auto-encoder model introduced in \citet{kipf2016variational} for unlabeled undirected graphs.
Our full link prediction model is schematically depicted in Figure~\ref{fig:model-c}.

In our experiments, we use the DistMult factorization~\cite{distmult-embedding_entities_and_relations} as the scoring function, which is known to perform well on standard link prediction benchmarks when used on its own.
In DistMult, every relation $r$ is associated with a diagonal matrix $R_r
\in \mathbb{R}^{d \times d}$ and a triple $(s, r, o)$ is scored as
\begin{equation}
f(s, r, o) = e_s^T R_r e_o \, .
\end{equation}

As in  previous work on factorization~\cite{distmult-embedding_entities_and_relations,complex-complex_embeddings_for_simple_link_prediction}, we train the model with negative sampling. For each observed example we sample $\omega$ negative ones. We sample by randomly corrupting either the subject or the object of each positive example.
We optimize for cross-entropy loss to push the model to score observable triples higher than the negative ones:
\begin{equation}
\begin{split}
\mathcal{L} = - \frac{1}{ (1+\omega) |\mathcal{\hat{E}}|}\sum\limits_{(s,r,o,y) \in \mathcal{T}} y \log l\bigl(f(s,r,o)\bigr) + \\
(1-y) \log\bigl(1-l\bigl(f(s,r,o)\bigr)\bigr) \, ,
\end{split}
\end{equation}
where $\mathcal{T}$ is the total set of real and corrupted triples, $l$ is the logistic sigmoid function, and $y$ is an indicator set to $y=1$ for positive triples and $y=0$ for negative ones.

\section{Empirical evaluation}

\subsection{Entity classification experiments}\label{sec:ent_exp}
Here, we consider the task of classifying entities in a knowledge base. In order to infer, for example, the type of an entity (e.g. person or company), a successful model needs to reason about the relations with other entities that this entity is involved in.

\paragraph{Datasets}
We evaluate our model on four datasets\footnote{http://dws.informatik.uni-mannheim.de/en/research/a-collection-of-benchmark-datasets-for-ml} in Resource Description Framework (RDF) format \cite{ristoski2016collection}: AIFB, MUTAG, BGS, and AM. Relations in these datasets need not necessarily encode directed subject-object relations, but are also used to encode the presence, or absence, of a specific feature for a given entity. In each dataset, the targets to be classified are properties of a group of entities represented as nodes. The exact statistics of the datasets can be found in Table \ref{table:classification_datasets}. For a more detailed description of the datasets the reader is referred to \citet{ristoski2016collection}. We remove relations that were used to create entity labels: \textit{employs} and \textit{affiliation} for AIFB, \textit{isMutagenic} for MUTAG, \textit{hasLithogenesis} for BGS, and \textit{objectCategory} and \textit{material} for AM.

\begin{table}[htp!]
\centering
\begin{tabular}{lrrrr}
\toprule
Dataset & AIFB & MUTAG & BGS & AM  \\ \midrule
Entities    & 8,285 & 23,644 & 333,845 & 1,666,764 \\
Relations   & 45 & 23 & 103 & 133 \\
Edges   & 29,043 & 74,227 & 916,199& 5,988,321 \\
Labeled  & 176 & 340 & 146 &  1,000 \\
Classes  & 4 & 2 & 2 &  11 \\

 \bottomrule
\end{tabular}
\caption{Number of entities, relations, edges and classes along with the number of labeled entities for each of the datasets. \textit{Labeled} denotes the subset of entities that have labels and that are to be classified.}
\label{table:classification_datasets}
\end{table}

\paragraph{Baselines}
As a baseline for our experiments, we compare against recent state-of-the-art classification results from RDF2Vec embeddings \cite{ristoski2016rdf2vec}, Weisfeiler-Lehman kernels (WL) \cite{shervashidze2011weisfeiler,de2015substructure}, and hand-designed feature extractors (Feat) \cite{paulheim2012unsupervised}. Feat assembles a feature vector from the in- and out-degree (per relation) of every labeled entity. RDF2Vec extracts walks on labeled graphs which are then processed using the Skipgram \cite{mikolov2013distributed} model to generate entity embeddings, used for subsequent classification. See \citet{ristoski2016rdf2vec} for an in-depth description and discussion of these baseline approaches. All entity classification experiments were run on CPU nodes with 64GB of memory.

\paragraph{Results}
All results in Table \ref{table:classification_results} are reported on the train/test benchmark splits from \citet{ristoski2016collection}. We further set aside 20\% of the training set as a validation set for hyperparameter tuning. For R-GCN, we report performance of a 2-layer model with 16 hidden units (10 for AM), basis function decomposition (Eq. \ref{eq:basis}), and trained with Adam \cite{kingma2014adam} for 50 epochs using a learning rate of $0.01$.  The normalization constant is chosen as $c_{i,r}=|\mathcal{N}_i^r|$. Further details on (baseline) models and hyperparameter choices are provided in the supplementary material.

\begin{table}[htp!]
\centering
\begin{tabular}{lcccc}
\toprule
Model & AIFB & MUTAG & BGS & AM  \\ \midrule
Feat & $55.55$ & $77.94$ & $72.41$ & $66.66$ \\
WL		& $80.55$ & $\mathbf{80.88}$ & $86.20$ & $87.37$  \\
RDF2Vec  & $88.88$ & $67.20$ & $\mathbf{87.24}$ & $88.33$ \\
R-GCN			& $\mathbf{95.83}$ & $73.23$ & $83.10$  & $\mathbf{89.29}$ \\\bottomrule

\end{tabular}
\caption{Entity classification results in accuracy (averaged over 10 runs) for a feature-based baseline (see main text for details), WL \cite{shervashidze2011weisfeiler,de2015substructure}, RDF2Vec \cite{ristoski2016rdf2vec}, and R-GCN (this work). Test performance is reported on the train/test set splits provided by \citet{ristoski2016collection}. \label{table:classification_results}} %R-GCN: Version of our model that uses a featurized version of the graph ('type' relations are transformed into node feature vectors). -- AIFB and AM are knowledge graphs, MUTAG are molecular graphs in RDF format, BGS is a feature table in RDF format. Note for R-GCN BGS (clean): same hyperparameters as in BGS.
\end{table}

Our model achieves state-of-the-art results on AIFB and AM. To explain the gap in performance on MUTAG and BGS it is important to understand the nature of these datasets. MUTAG is a dataset of molecular graphs, which was later converted to RDF format, where relations either indicate atomic bonds or merely the presence of a certain feature. BGS is a dataset of rock types with hierarchical feature descriptions which was similarly converted to RDF format, where relations encode the presence of a certain feature or feature hierarchy. Labeled entities in MUTAG and BGS are only connected via high-degree hub nodes that encode a certain feature.

We conjecture that the fixed choice of normalization constant for the aggregation of messages from neighboring nodes is partly to blame for this behavior, which can be particularly problematic for nodes of high degree. A potential way to overcome this limitation is to introduce an attention mechanism, i.e. to replace the normalization constant $1/c_{i,r}$ with data-dependent attention weights $a_{ij,r}$, where $\sum_{j,r}a_{ij,r}=1$. We expect this to be a promising avenue for future research.

\subsection{Link prediction experiments}
As shown in the previous section, R-GCNs serve as an effective encoder for relational data. We now combine our encoder model with a scoring function (which we will refer to as a decoder, see Figure \ref{fig:model-c}) to score candidate triples for link prediction in knowledge bases.

\subsubsection{Datasets}
Link prediction algorithms are commonly evaluated on FB15k, a subset of the relational database Freebase, and WN18, a subset of WordNet containing lexical relations between words. In \citet{toutanova2015observed}, a serious flaw was observed in both datasets: The presence of inverse triplet pairs $t = (e_1, r, e_2)$ and $t' = (e_2, r^{-1}, e_1)$ with $t$ in the training set and $t'$ in the test set. This reduces a large part of the prediction task to memorization of affected triplet pairs. A simple baseline LinkFeat employing a linear classifier on top of sparse feature vectors of observed training relations was shown to outperform existing systems by a large margin. To address this issue, Toutanova and Chen proposed a reduced dataset FB15k-237 with all such inverse triplet pairs removed. We therefore choose FB15k-237 as our primary evaluation dataset. Since FB15k and WN18 are still widely used, we also include results on these datasets using the splits introduced by \citet{bordes2013translating}.

\begin{table}[ht]
\centering
\begin{tabular}{lrrr}
\toprule
Dataset & WN18 & FB15K & FB15k-237\\ \midrule
Entities    & 40,943  & 14,951 & 14,541\\
Relations   & 18 & 1,345 & 237\\
Train edges & 141,442 & 483,142 & 272,115\\
Val.~edges & 5,000 & 50,000 & 17,535\\
Test edges & 5,000 & 59,071 & 20,466\\ \bottomrule
\end{tabular}
\caption{Number of entities and relation types along with the number of edges per split for the three datasets.\label{table:datasets}}
\end{table}

\paragraph{Baselines}
A common baseline for both experiments is direct optimization of \textit{DistMult}~\cite{distmult-embedding_entities_and_relations}. This factorization strategy is known to perform well on standard datasets, and furthermore corresponds to a version of our model with fixed entity embeddings in place of the R-GCN encoder as described in Section \ref{section:link_prediction}. As a second baseline, we add the simple neighbor-based LinkFeat algorithm proposed in \citet{toutanova2015observed}.

We further compare to ComplEx \cite{complex-complex_embeddings_for_simple_link_prediction} and HolE \cite{nickel2015holographic}, two state-of-the-art link prediction models for FB15k and WN18. ComplEx facilitates modeling of asymmetric relations by generalizing DistMult to the complex domain, while HolE replaces the vector-matrix product with circular correlation. Finally, we include comparisons with two classic algorithms -- CP~\cite{hitchcock1927expression} and TransE~\cite{bordes2013translating}.

\begin{table*}[t!]
\centering
\begin{tabular}{@{}lllllllllll@{}}
\toprule
      & \multicolumn{5}{c}{FB15k}                             & \multicolumn{5}{c}{WN18}                            \\ \cmidrule(lr){2-6} \cmidrule(l){7-11}
      & \multicolumn{2}{c}{MRR} & \multicolumn{3}{c}{Hits @} & \multicolumn{2}{c}{MRR} & \multicolumn{3}{c}{Hits @} \\ \cmidrule(lr){2-3} \cmidrule(lr){4-6} \cmidrule(lr){7-8} \cmidrule(l){9-11}
Model & Raw      & Filtered     & 1       & 3      & 10      & Raw      & Filtered     & 1       & 3      & 10      \\ \midrule
LinkFeat 		&  & 0.779 & & & 0.804 & & 0.938 & & & 0.939 \\
\midrule
DistMult 		& 0.248 & 0.634 & 0.522 & 0.718 & 0.814 & 0.526 & 0.813 & 0.701 & 0.921 & 0.943 \\
R-GCN  			& 0.251 & 0.651 & 0.541 & 0.736 & 0.825 & 0.553 & 0.814 & 0.686 & 0.928 & 0.955 \\
R-GCN+			& \textbf{0.262} & \textbf{0.696} & \textbf{0.601} & \textbf{0.760} & \textbf{0.842} & 0.561 & 0.819 & 0.697 & 0.929 & \textbf{0.964}  \\ \midrule
CP* & 0.152 & 0.326 & 0.219 & 0.376 & 0.532 & 0.075 & 0.058 & 0.049 & 0.080 & 0.125 \\
TransE* 		& 0.221 & 0.380 & 0.231 & 0.472 & 0.641 & 0.335 & 0.454 & 0.089 & 0.823 & 0.934 \\
HolE** 			& 0.232 & 0.524 & 0.402 & 0.613 & 0.739 & \textbf{0.616} & 0.938 & 0.930 & \textbf{0.945} & 0.949\\
ComplEx* 		& 0.242 & 0.692 & 0.599 & 0.759 & 0.840 & 0.587 & \textbf{0.941} & \textbf{0.936} & \textbf{0.945} & 0.947 \\ \bottomrule
\end{tabular}
\caption{Results on the the Freebase and WordNet datasets. Results marked (*) taken from \citet{complex-complex_embeddings_for_simple_link_prediction}. Results marks (**) taken from \citet{nickel2015holographic}. R-GCN+ denotes an ensemble between R-GCN and DistMult -- see main text for details.}
\label{table:link-results}
\end{table*}

\paragraph{Results}
We provide results using two commonly used evaluation metrics: \textit{mean reciprocal rank} (MRR) and \textit{Hits at $n$} (H@n).
Following \citet{bordes2013translating}, both metrics can be computed in a raw and a filtered setting. We report both filtered and raw MRR (with filtered MRR typically considered more reliable), and filtered Hits at 1, 3, and 10.

We evaluate hyperparameter choices on the respective validation splits. We found a normalization constant defined as $c_{i,r}=c_{i}=\sum_{r} |\mathcal{N}^r_i|$ --- in other words, applied across relation types -- to work best. For FB15k and WN18, we report results using basis decomposition (Eq. \ref{eq:basis}) with two basis functions, and a single encoding layer with $200$-dimensional embeddings. For FB15k-237, we found block decomposition (Eq. \ref{eq:block}) to perform best, using two layers with block dimension $5 \times 5$ and $500$-dimensional embeddings. We regularize the encoder through edge dropout applied before normalization, with dropout rate $0.2$ for self-loops and $0.4$ for other edges. Using edge droupout makes our training objective similar to that of denoising autoencoders~\cite{vincent2008}. We apply $l2$ regularization to the decoder with a penalty of $0.01$.

We use the Adam optimizer \cite{kingma2014adam} with a learning rate of $0.01$. For the baseline and the other factorizations, we found the parameters from \citet{complex-complex_embeddings_for_simple_link_prediction} -- apart from the dimensionality on FB15k-237 -- to work best, though to make the systems comparable we maintain the same number of negative samples (i.e.~$\omega=1$). We use full-batch optimization for both the baselines and our model.

\begin{figure}
\centering
\begin{tikzpicture}
\begin{axis}[xlabel=Average degree,
ylabel=MRR,
legend style={font=\fontsize{9}{11}\selectfont},
width=1.2\linewidth,
height=0.6\linewidth,
scale=.65]
\addplot[color=red,mark=triangle*] table[x=Degree,y=GCN] {data/degree.dat};
\label{p1}
\addplot[color=blue,mark=x] table[x=Degree, y=Distmult] {data/degree.dat};
\label{p2}
\legend{R-GCN, DistMult}
\end{axis}%
\end{tikzpicture}\vspace{-0.5em}
\caption{Mean reciprocal rank (MRR) for R-GCN  and DistMult on the FB15k validation data as a function of the node degree (average of subject and object).\label{figure:degree}}
\end{figure}
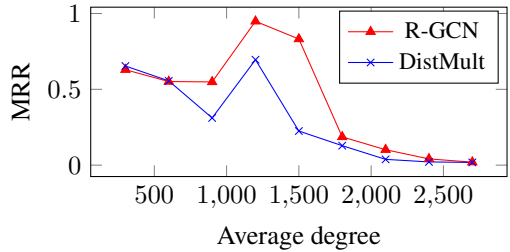

On FB15k, local context in the form of inverse relations is expected to dominate the performance of the factorizations, contrasting with the design of the R-GCN model. To better understand the difference, we plot in Figure \ref{figure:degree} the FB15k performance of the best R-GCN model and the baseline (DistMult) as functions of degree of nodes corresponding to entities in the considered triple
(namely, the average of degrees for the subject and object entities). It can be seen that our model performs better for nodes with high degree where contextual information is abundant. The observation that the two models are complementary suggests combining the strengths of both into a single model, which we refer to as R-GCN+. On FB15k and WN18 where local and long-distance information can both provide strong solutions, we expect R-GCN+ to outperform each individual model. On FB15k-237 where local information is less salient, we do not expect the combination model to outperform a pure R-GCN model significantly. To test this, we evaluate an ensemble (R-GCN+) with a trained R-GCN model and a separately trained DistMult factorization model: $f(s,r,t)_{\text{R-GCN+}} = \alpha f(s,r,t)_{\text{R-GCN}} + (1- \alpha) f(s,r,t)_{\text{DistMult}}$, with $\alpha=0.4$ selected on FB15k development data.

In Table~\ref{table:link-results}, we evaluate the R-GCN model and the combination model (R-GCN+) on FB15k and WN18.

On the FB15k and WN18 datasets, R-GCN and R-GCN+ both outperform the DistMult baseline, but like all other systems underperform on these two datasets compared to the LinkFeat algorithm. The strong result from this baseline highlights the contribution of inverse relation pairs to high-performance solutions on these datasets. Interestingly, R-GCN+ yields better performance than ComplEx for FB15k, even though the R-GCN decoder (DistMult) does not explicitly model asymmetry in relations, as opposed to ComplEx.

This suggests that combining the R-GCN encoder with the ComplEx scoring function (decoder) may be a promising direction for future work. The choice of scoring function is orthogonal to the choice of encoder; in principle, any scoring function or factorization model could be incorporated as a decoder in our auto-encoder framework.

\begin{table}[htp!]
\centering
\begin{tabular}{llllll}
\toprule
                                  & \multicolumn{2}{c}{MRR} & \multicolumn{3}{c}{Hits @} \\ \cmidrule{2-3} \cmidrule{4-6}
Model                             & Raw      & Filtered     & 1       & 3       & 10     \\ \midrule
LinkFeat     &  & 0.063         &    & & 0.079  \\ \midrule
DistMult     & 0.100         & 0.191         & 0.106   & 0.207   & 0.376  \\
R-GCN               & \textbf{0.158}       & 0.248 & \textbf{0.153}     & 0.258     & 0.414     \\
R-GCN+                     & 0.156 & \textbf{0.249} & 0.151     & \textbf{0.264}     & \textbf{0.417} \\     \midrule
CP     & 0.080 & 0.182 & 0.101 & 0.197 & 0.357 \\
TransE    	& 0.144	& 0.233 & 0.147 & 0.263 &	0.398  \\
HolE    	& 0.124	& 0.222 & 0.133 & 0.253 &	0.391  \\
ComplEx    & 0.109 & 0.201 & 0.112 & 0.213   & 0.388 \\
\bottomrule
\end{tabular}
\caption{Results on FB15k-237, a reduced version of FB15k with problematic inverse relation pairs removed. CP, TransE, and ComplEx were evaluated using the code published for \citet{complex-complex_embeddings_for_simple_link_prediction}, while HolE was evaluated using the code published for \citet{nickel2015holographic}.}
\label{table:fb15k-237}
\end{table}

In Table \ref{table:fb15k-237}, we show results for FB15k-237 where (as previously discussed) inverse relation pairs have been removed and the LinkFeat baseline fails to generalize\footnote{Our numbers are not directly comparable to those reported in \citet{toutanova2015observed}, as they use pruning both for training and testing (see their sections 3.3.1 and 4.2). Since their pruning schema is not fully specified (e.g., values of the relation-specific parameter $t$ are not given) and the code is not available, it is not possible to replicate their set-up.}. Here, our R-GCN model outperforms the DistMult baseline by a large margin of 29.8\%, highlighting the importance of a separate encoder model. As expected from our earlier analysis, R-GCN and R-GCN+ show similar performance on this dataset. The R-GCN model further compares favorably against other factorization methods, despite relying on a DistMult decoder which shows comparatively weak performance when used without an encoder.

\section{Related Work}
\subsection{Relational modeling}
Our encoder-decoder approach to link prediction relies on DistMult~\cite{distmult-embedding_entities_and_relations} in the decoder,  a special and simpler case of the RESCAL factorization~\cite{nickel2011three}, more effective than the original RESCAL in the context of multi-relational knowledge bases. Numerous alternative factorizations have been proposed and studied in the context of SRL, including both (bi-)linear and nonlinear ones~(e.g., \cite{bordes2013translating,socher2013reasoning,KaiWei14,nickel2015holographic,complex-complex_embeddings_for_simple_link_prediction}). Many of these approaches can be regarded as modifications or special cases of classic tensor decomposition methods such as CP or Tucker; for a comprehensive overview of tensor decomposition literature we refer the reader to \citet{kolda2009tensor}.

Incorporation of paths between entities in knowledge
bases has recently received considerable attention. We can roughly classify previous work into (1) methods creating auxiliary triples, which are then added to the learning objective of a factorization model~\cite{guu2015traversing,garcia2015composing}; (2) approaches using paths (or walks) as features when predicting edges ~\cite{lin2015modeling}; or (3) doing both at the same time~\cite{neelakantan2015compositional,toutanova2016compositional}. The first direction is largely orthogonal to ours, as we would also expect improvements from adding similar terms to our loss (in other words, extending our decoder). The second research line is more comparable; R-GCNs provide a computationally cheaper alternative to these path-based models. %thus, letting us use larger datasets than typically considered with these methods.
Direct comparison is somewhat complicated as path-based methods used different datasets (e.g., sub-sampled sets of walks from a knowledge base).

\subsection{Neural networks on graphs}
Our R-GCN encoder model is closely related to a number of works in the area of neural networks on graphs. It is primarily motivated as an adaption of previous work on GCNs \cite{bruna2014spectral,duvenaud2015convolutional,defferrard2016convolutional,kipf2016semi} for large-scale and highly multi-relational data, characteristic of realistic knowledge bases.

Early work in this area includes the graph neural network by \citet{scarselli2009graph}. A number of extensions to the original graph neural network have been proposed, most notably \cite{li2015gated} and \cite{pham2016column}, both of which utilize gating mechanisms to facilitate optimization.

R-GCNs can further be seen as a sub-class of message passing neural networks \cite{gilmer2017neural}, which encompass a number of previous neural models for graphs, including GCNs, under a differentiable message passing interpretation.

\section{Conclusions}

We have introduced relational graph convolutional networks (R-GCNs) and demonstrated their effectiveness in the context of two standard statistical relation modeling problems: link prediction and entity classification. For the entity classification problem, we have demonstrated that the R-GCN model can act as a competitive, end-to-end trainable graph-based encoder. For link prediction, the R-GCN model with DistMult factorization as the decoding component outperformed direct optimization of the factorization model, and achieved competitive results on standard link prediction benchmarks. Enriching the factorization model with an R-GCN encoder proved especially valuable for the challenging FB15k-237 dataset, yielding a 29.8\% improvement over the decoder-only baseline.

There are several ways in which our work could be extended. For example, the graph autoencoder model could be considered in combination with other factorization models, such as ComplEx~\cite{complex-complex_embeddings_for_simple_link_prediction}, which can be better suited for modeling asymmetric relations. It is also straightforward to integrate entity features in R-GCNs, which would be beneficial both for link prediction and entity classification problems. To address the scalability of our method, it would be worthwhile to explore subsampling techniques, such as in \citet{hamilton2017inductive}. Lastly, it would be promising to replace the current form of summation over neighboring nodes and relation types with a data-dependent attention mechanism. Beyond modeling knowledge bases, R-GCNs can be generalized to other applications where relation factorization models have been shown effective (e.g. relation extraction).

\subsection*{Acknowledgements}
We would like to thank Diego Marcheggiani, Ethan Fetaya, and Christos Louizos for helpful discussions and comments.
This project is supported by the European Research Council (ERC StG BroadSem 678254), the SAP Innovation
Center Network
and the Dutch National Science Foundation (NWO VIDI 639.022.518).

\bibliographystyle{aaai}
\bibliography{bibliography}

\newpage
\onecolumn
\setlength{\parskip}{0.5em}
\appendix
\section*{Further experimental details on entity classification}
\vspace{2em}

For the entity classification benchmarks described in our paper, the evaluation process differs subtly between publications. To eliminate these differences, we repeated the baselines in a uniform manner, using the canonical test/train split from \cite{ristoski2016collection}. We performed hyperparameter optimization on only the training set, running a single evaluation on the test set after hyperparameters were chosen for each baseline. This explains why the numbers we report differ slightly from those in the original publications (where cross-validation accuracy was reported).

For WL, we use the \textit{tree} variant of the Weisfeiler-Lehman subtree kernel from the Mustard library.\footnote{https://github.com/Data2Semantics/mustard}
For RDF2Vec, we use an implementation provided by the authors of \cite{ristoski2016rdf2vec} which builds on Mustard. In both cases, we extract explicit feature vectors for the instance nodes, which are classified by a linear SVM.

For the MUTAG task, our preprocessing differs from that used in \cite{de2015substructure,ristoski2016rdf2vec} where for a given target relation $(s, r, o)$ all triples connecting $s$ to $o$ are removed. Since $o$ is a boolean value in the MUTAG data, one can infer the label after processing from other boolean relations that are still present. This issue is now mentioned in the Mustard documentation. In our preprocessing, we remove only the specific triples encoding the target relation.

Hyperparameters for baselines are chosen according to the best model performance in \cite{ristoski2016rdf2vec}, i.e.~WL: 2 (tree depth), 3 (number of iterations); RDF2Vec: 2 (WL tree depth), 4 (WL iterations), 500 (embedding size), 5 (window size), 10 (SkipGram iterations), 25 (number of negative samples). We optimize the SVM regularization constant $C\in\{0.001, 0.01, 0.1, 1, 10, 100, 1000\}$ based on performance on a 80/20 train/validation split (of the original training set).

For R-GCN, we choose an $l2$ penalty on first layer weights $C_{l2}\in\{0, 5\cdot10^{-4}\}$ and the number of basis functions $B\in\{0, 10, 20, 30, 40\}$ based on validation set performance, where $B=0$ refers to using no basis function decomposition. Using the block decomposition did not improve results. Otherwise, hyperparameters are chosen as follows: $50$ (number of epochs), $16$ (number of hidden units), and $c_{i,r}=|\mathcal{N}^r_i|$ (normalization constant). We do not use dropout. For AM, we use a reduced number of $10$ hidden units for R-GCN to reduce the memory footprint.

Results with standard error (omitted in main paper due to spatial constraints) are summarized in Table \ref{table:results}. All entity classification experiments were run on CPU nodes with 64GB of memory.

\begin{table*}[htp!]
\centering
\begin{tabular}{lrrrr}
\toprule
R-GCN setting & AIFB & MUTAG & BGS & AM  \\ \midrule
$l2$ penalty & $0$ & $5\cdot 10^{-4}$ & $5\cdot 10^{-4}$ & $5\cdot 10^{-4}$ \\
\# basis functions & $0$ & $30$ & $40$ & $40$  \\
\# hidden units & $16$ & $16$ & $16$ & $10$ \\ \bottomrule
\end{tabular}
\caption{Best hyperparameter choices based on validation set performance for 2-layer R-GCN model. \label{table:hyperparams}}
\end{table*}

\begin{table*}[htp!]
\centering
\begin{tabular}{lcccc}
\toprule
Model & AIFB & MUTAG & BGS & AM  \\ \midrule
Feat & $55.55\pm0.00$ & $77.94 \pm 0.00$ & $72.41\pm0.00$ & $66.66 \pm 0.00$ \\
WL		& $80.55\pm0.00$ & $\mathbf{80.88}\pm 0.00$ & $86.20\pm 0.00$ & $87.37 \pm 0.00$  \\
RDF2Vec  & $88.88 \pm 0.00$ & $67.20 \pm 1.24$ & $\mathbf{87.24}  \pm 0.89$ & $88.33 \pm 0.61$ \\
R-GCN (Ours)			& $\mathbf{95.83} \pm 0.62$ & $73.23 \pm 0.48$ & $83.10\pm 0.80$  & $\mathbf{89.29} \pm 0.35$ \\\bottomrule
\end{tabular}
\caption{Entity classification results in accuracy (average and standard error over 10 runs) for a feature-based baseline (see main text for details), WL \cite{shervashidze2011weisfeiler,de2015substructure}, RDF2Vec \cite{ristoski2016rdf2vec}, and R-GCN (this work). Test performance is reported on the train/test set splits provided by \cite{ristoski2016collection}. \label{table:results}}
\end{table*}

\end{document}